\documentclass{bmvc2k}
\usepackage{multirow}
\usepackage{amssymb}
\usepackage{booktabs}
\title{Neighbor Regularized Bayesian Optimization for Hyperparameter Optimization}

\addauthor{Lei Cui}{cuil19@mails.tsinghua.edu.cn}{1}
\addauthor{Yangguang Li}{liyangguang@sensetime.com}{2}
\addauthor{Xin Lu}{luxin@sensetime.com}{2}
\addauthor{Dong An}{dong.an@cripac.ia.ac.cn}{3}
\addauthor{Fenggang Liu}{liufgtech@163.com}{2}

\addinstitution{
 Tsinghua University \\
 Beijing, China
}
\addinstitution{
 SenseTime Research \\
 Beijing, China
}
\addinstitution{
 Institute of Automation, Chinese Academy of Sciences \\
 Beijing, China
}

\runninghead{Lei, et al.}{Neighbor Regularized Bayesian Optimization}

\begin{document}

\maketitle

\begin{abstract}
\textit{Bayesian Optimization} (BO) is a common solution to search optimal hyperparameters based on sample observations of a machine learning model. Existing BO algorithms could converge slowly even collapse when the potential observation noise misdirects the optimization. In this paper, we propose a novel BO algorithm called \textit{Neighbor Regularized Bayesian Optimization} (NRBO) to solve the problem. We first propose a neighbor-based regularization to smooth each sample observation, which could reduce the observation noise efficiently without any extra training cost. Since the neighbor regularization highly depends on the sample density of a neighbor area, we further design a density-based acquisition function to adjust the acquisition reward and obtain more stable statistics. In addition, we design a adjustment mechanism to ensure the framework maintains a reasonable regularization strength and density reward conditioned on remaining computation resources. We conduct experiments on the bayesmark benchmark and important computer vision benchmarks such as ImageNet and COCO. Extensive experiments demonstrate the effectiveness of NRBO and it consistently outperforms other state-of-the-art methods.
\end{abstract}

%-------------------------------------------------------------------------
\section{Introduction}
The performance of modern machine learning models highly depends on the good choice of hyperparameters. Thus a system that can quickly and automatically optimize the hyperparameters becomes more and more important nowadays.
The hyperparameter optimization problem is often regarded as a black-box optimization problem, and the common solution is \textit{Bayesian optimization} (BO)~\cite{golovin2017google,wu2019hyperparameter,eriksson2019scalable,2022smac3,2021hpobench,2021MOBO,victoria2021automatic}.

Though existing BO algorithms have shown significant improvement compared with the random search algorithm~\cite{li2017hyperband,falkner2018bohb}, BO could still converge slowly even collapse when the potential observation noise is non-negligible.
Specifically, the observation (a.k.a. performance) of a model could fluctuate in a wide range even using the same hyperparameters, which might mislead the surrogate model to overfit noisy observations.
A natural idea to solve the problem is to obtain reliable observations by training repetitively. 
However, it will cause multiplied computational cost which is unacceptable in heavy tasks. 
Another idea is to add noise assumptions on the sample observations, while the noise assumptions without any prior will decrease the fitting efficiency of the surrogate model.

To overcome the above difficulties, we propose a novel BO algorithm called \textit{Neighbor Regularized Bayesian Optimization} (NRBO). 
NRBO applies adaptive regularization at different positions of the search space according to the neighbor sample points.  
Specifically, the observation of a sample will be smoothed by the observations of its neighbor samples. Then, the smoothed observations are used to optimize the surrogate model.
The regularization could reduce the observation noise \textit{which has the same spirit as the k-nearest neighbors algorithm}~\cite{fix1989knn}. Thus, no repetitive training phases are needed to get extra observations. 
Compared with directly introducing a global noise assumption, this method can dynamically smooth observations at different regions thus regularize the optimization.

The neighbor-based regularization algorithm needs to have as many samples as possible at each desired area to obtain credible and stable statistics. 
Therefore, we design a new density-based acquisition function, which can adjust the acquisition reward according to the sample density of a certain area. 
The proposed acquisition function adopts a stronger reward for the \textit{sparse} area where the density of observed samples is low. 
This design has two advantages. 
First, it provides a more balanced sampling strategy that the observed samples are distributed more evenly on the entire search space.
Second, it prevents the surrogate model from sticking into a local sub-optimal solution prematurely.

Finally, we design a mechanism to dynamically adjust the regularization strength and acquisition reward. 
This mechanism ensures that the framework maintains a reasonable regularization strength and acquisition reward during the Bayesian optimization process.
The regularization strength and density reward are positively correlated with the remaining computation resources. 
As the search progresses, the regularization strength will be weakened to help the model fit more subtle data patterns in finer spatial resolution. 
At the same time, the density reward will also be weakened to balance the exploration-exploitation trade-off and encourage the algorithm to converge at the high-performance area when searching is about to end.

Extensive experiments demonstrate the effectiveness of our NRBO. 
NRBO achieves the state-of-the-art performance on the bayesmark and six commonly used computer vision tasks.
For example, NRBO achieves 0.85\%, 0.22\%, 2.1\% and 2.36\% higher accuracy on Stanfordcar, ImageNet, VOC and COCO datasets compared with random search~\cite{bergstra2012random}.
We summarize our main contributions in this paper as follows:

$\bullet$ We propose a novel Bayesian optimization method based on the statistics of the neighbor observations, which can adaptively reduce the observation noise and regularize the surrogate model.

$\bullet$ We propose a novel density-based acquisition function to provide credible and stable statistics for Bayesian optimization process.

$\bullet$ We design a mechanism to dynamically adjust the regularization strength and acquisition reward according to the remaining computation resource. It helps to fit more subtle data patterns and balance the exploration-exploitation trade-off as the optimization progresses.

%--------------------------------------------------------------------------------------------------------------------------------------------------
\section{Related Works}
% In this section, we briefly introduce the mainstream solutions for hyper-parameter optimization. 
%--------------------------------------------------------------------------------------------------------------------------------------------------
\subsection{Model-free Method}
The most straightforward approach to optimize the hyperparameters is the model-free method. 
Grid search is one of the commonly used methods that can be easily implemented and parallelized~\cite{lerman1980fitting, liashchynskyi2019grid}. 
It discretizes the search space into a mesh grid and evaluates them all. 
The computation cost will explode exponentially as the dimension or the resolution of the hyperparameter increases. 
Another model-free approach is the Random search~\cite{bergstra2012random,solis1981minimization}. 
Instead of traversing the discretized search space,  random search selects the candidate hyperparameter randomly. 
Although the  Random Search and the Grid Search seems to have similar efficiency,  Random Search usually performs better in a limited search budget in practice. 
The reason is that the model performance is usually not distributed uniformly in the entire search space. 
The Random Search samples a fixed number of parameter combinations from the specified distribution, which improves system efficiency by reducing the probability of wasting much time on a small poorly-performing region~\cite{yang2020hyperparameter}.

%--------------------------------------------------------------------------------------------------------------------------------------------------
\subsection{Bayesian Optimization}
In addition to the model-free approaches, the most popular method used in hyperparameter optimization is Bayesian optimization~\cite{golovin2017google,wu2019hyperparameter,eriksson2019scalable,2021MOBO,victoria2021automatic,shahriari2015taking,snoek2015scalable,swersky2013multi}. 
Instead of searching the hyperparameter in a pre-defined distribution, Bayesian optimization can dynamically fit the observed data to determine the next search place. 
Bayesian optimization algorithm runs a search loop that iteratively fits the surrogate model and queries the acquisition function~\cite{snoek2012practical}. 
By fitting the observed data, the surrogate model can predict the performance distribution throughout the search space with prior distribution such as smoothness.
The acquisition function selects the promising regions that balance the exploration-exploitation trade-off~\cite{hazan2017hyperparameter}. 
A typical Bayesian optimization algorithm is the Gaussian Process Bayesian Optimization (GPBO)~\cite{rasmussen2003gaussian}. GPBO utilizes the Gaussian process as the surrogate model that can predict the performance in the search space as a jointly Gaussian distribution. 
SMAC is another popular Bayesian optimization algorithm that uses random forest as the surrogate model and ensembled regression trees as the objective function~\cite{hutter2011sequential}. 
HEBO enhances the surrogate model through input warping and output transformations and proposes a multi-objective acquisition function for candidate selection~\cite{cowen2020hebo}.

%--------------------------------------------------------------------------------------------------------------------------------------------------
\section{Method}

% We first briefly introduce the concepts and pipeline of a typical Bayesian optimization in section~\ref{subsec:3.1}. Then we detail our proposed \textit{Neighbor Regularized Bayesian Optimization} (NRBO) in section~\ref{subsec:3.2}.

%--------------------------------------------------------------------------------------------------------------------------------------------------
\subsection{Preliminaries}\label{subsec:3.1}

Hyperparameter optimization can be seen as a black-box optimization problem. The searching process is started with a initial observation dataset $\mathcal{D}_{0}$. Specifically, a batch of random hyperparameters $\boldsymbol{x}_{0:m-1}$ are sampled at first. Then models with these hyperparameters are trained and evaluated to get the observations $\boldsymbol{y}_{0:m-1}$. The observation dataset $\mathcal{D}_{0}$ is initialized by $\{(\boldsymbol{x}_{0:m-1},\boldsymbol{y}_{0:m-1})\}$. The surrogate model $f_\theta$ will be trained for $N$ times to fit the observation dataset $\mathcal{D}_{0}$, where $\theta$ is optimized by minimizing the negative log marginal likelihood.

After fitting the surrogate model, we need to guess the best $\boldsymbol{x^*}$ by proposing a new (batch of) sample point according to acquisition function $f_{acq}$. Given an input sample point $\boldsymbol{x}$, $f_{acq}$ calculates the priority of this sample point according to probability distribution predicted by the surrogate model $f_{\theta}$. The acquisition function is usually designed to balance the exploration-exploitation trade-off.  A novel sample point $\boldsymbol{\hat{x}}$ is selected by $argmax_{\boldsymbol{x} \in \mathcal{X}}f_{acq}(\boldsymbol{x})$, and in practice we discretize the continuous search space $\mathcal{X}$ into a meshgrid points collection. 

Once the novel sample point is selected, it will be evaluated on black-box by training the machine learning models with the hyperparameter it represents. Then the observation $\{(\boldsymbol{\hat{x}},\hat{y})\}$ will be added to datasets $\mathcal{D}_{i}$ for the next iteration. When the main loop is over, the algorithm returns the best-performing data point as the result.

%--------------------------------------------------------------------------------------------------------------------------------------------------
\begin{figure*}[htbp]
\centering
    % \begin{center}
       \includegraphics[width=0.7\textwidth]{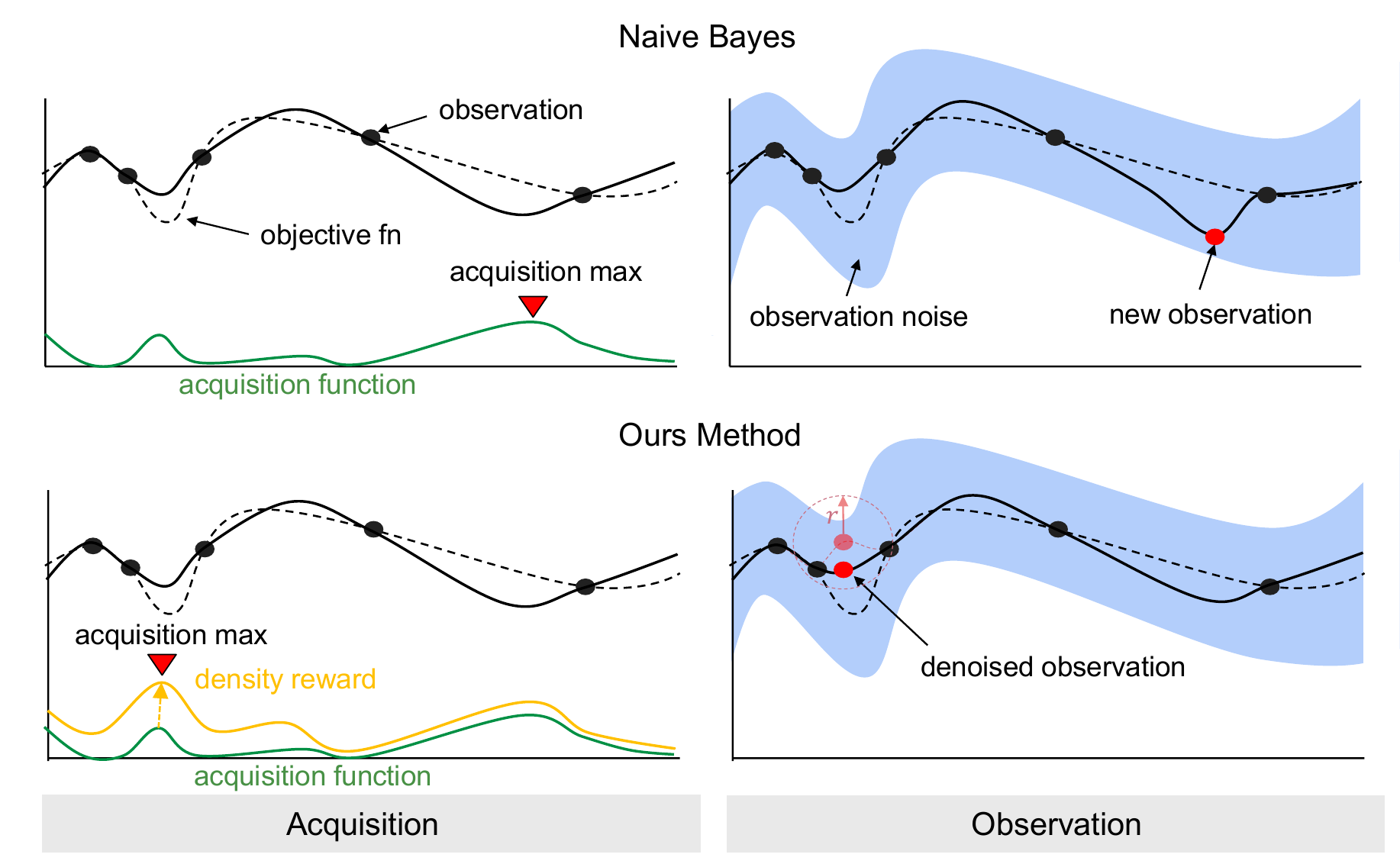}
    % \end{center}
   \caption{Illustration of the NRBO algorithm. In two key progress of acquisition and observation. Our method is different from the naive bayes algorithm. In acquisition stage, we propose a density-based acquisition function to accelerate the acquisition process, in which  adjacent sample points in the neighbor are considered. In observation stage, neighbor regularized mechanism is introduced to smooth the observation noise and release the burden of repetitive observation.}
   \label{fig:brief}
\end{figure*}
%--------------------------------------------------------------------------------------------------------------------------------------------------

\subsection{Neighbor Regularized Bayesian Optimization}\label{subsec:3.2}

Figure~\ref{fig:brief} shows a brief illustration of our NRBO. The main innovations of NRBO can be summarized as: a novel regularization to reduce the observation noise (section~\ref{subsec:3.2.1}), a density-based acquisition function to adjust the acquisition reward (section~\ref{subsec:3.2.2}) and a dynamic adjustment of the regularization strength and acquisition reward according to the remaining computation resource (section~\ref{subsec:3.2.3}). 

%--------------------------------------------------------------------------------------------------------------------------------------------------
\subsubsection{\textbf{Neighbor-based regularization}}\label{subsec:3.2.1}

In the context of Bayesian optimization, the algorithm maintains an observation dataset $\mathcal{D}_{i}$ at any iteration $i$, which consists of all sampled hyperparameters as well as their corresponding observation received from the black box.
To regularize the surrogate model, we define 
\begin{equation}\label{eq3}
    \overline{y_j} = \frac{\sum_{k}^{N_D} P_j(\boldsymbol{x}_k) \cdot y_k}{\sum_{k}^{N_D} P_j(\boldsymbol{x}_k)}, \\
    \quad P_j(\boldsymbol{x}_k) = \left\{ 
    \begin{array}{ll}
    1 & \qquad\Vert \boldsymbol{x}_j - \boldsymbol{x}_k \Vert \leq \sigma_1 \\
    0 & \qquad\Vert \boldsymbol{x}_j - \boldsymbol{x}_k \Vert > \sigma_1 
    \end{array} 
    \right.   
\end{equation}

where $N_D$ is the number of samples in $\mathcal{D}_{i}$, $P_j(\boldsymbol{x}_k)$ is a filter function that keeps the points in the neighbor of $\boldsymbol{x}_j$ with $\sigma_1$ as radius. 
$\overline{y_j}$ is the observation smoothed by its neighbor sample observations, and we have a smoothed observation dataset $\overline{\mathcal{D}_{i}} = \{(\boldsymbol{x}_l,\overline{y_l})\}_{l=0:N-1}$. Different from naive Bayes, the surrogate model is regularized by fitting the $\overline{\mathcal{D}_{i}}$ instead of ${\mathcal{D}_{i}}$.

Here we give a brief analysis on neighbor-based regularization. 
Firstly, NRBO smoothes each observation by taking the neighbor observations into account.
In this way, NRBO can efficiently regularize the model without repetitive training. 
In practice, the computation burden of repetitive training is unacceptable because each observation  needs to re-train the entire model. 
Secondly, the strength of regularization is determined by the statistical properties of all sample points lie in the neighbor area. As the variance of observations increase, NRBO will also impose stronger regularization. Compared with simply modeling noise with $y=f(x)+\eta$ that regularize the model evenly throughout the entire search space, the NRBO regularization mechanism is more adaptive and effective. Note that in practice, $\sigma_1$ moves according to the searching progress, we will discuss it in section~\ref{subsec:3.2.3}.

%--------------------------------------------------------------------------------------------------------------------------------------------------
\subsubsection{\textbf{Density-based acquisition function}}\label{subsec:3.2.2}
As demonstrated in section \ref{subsec:3.2.1}, NRBO imposes adaptive regularization by adopting neighbor sample observations. Obviously, to make the NRBO works more efficiently, we need to increase the density of the sample points for a more reliable and stable statistics. 
If there exists no other observed sample point lies in the neighbor area of a certain point, NRBO degenerates into a normal Bayesian optimization without any regularization in this area.

Based on the above analysis, we propose a density-based strategy to adjust the acquisition function. Our baseline acquisition function is the multi-objective acquisition ensemble (MACE) proposed in HEBO~\cite{cowen2020hebo}. The proposed density-based acquisition function can be formally denoted as:
\begin{equation}\label{eq5}
     min (-f_{EI}-g_{d}(\boldsymbol{x})S_{EI};\ -f_{PI}-g_{d}(\boldsymbol{x})S_{PI};\ f_{UCB}-g_{d}(\boldsymbol{x})S_{UCB}), \quad g_{d}(\boldsymbol{x}) = e^{-f_n(\boldsymbol{x},\sigma_2)}  \\
\end{equation}
where $f_{EI}$, $f_{PI}$ and $f_{UCB}$ are three widely-used myopic acquisition functions. $S_{EI}$, $S_{PI}$ and $S_{UCB}$ are norm items that represent the standard deviation of acquisition values of all candidate points in search space. We follow HEBO~\cite{cowen2020hebo} to define and use these symbols, please refer to HEBO for more details. Different from HEBO, we introduce new adjustment items $g_d$. In equation \ref{eq5}, function $f_n$ returns the number of sample points in $D_i$ that lies in the neighbor area of point $\boldsymbol{x}$. $\sigma_2$ is the radius that defines the neighbor range.

With adjustment item $g_d$, the acquisition value of the candidate point $\boldsymbol{x}$ can be adjusted by the density of its observed adjacent sample points in $D_{i}$, thus $g_d$ can be regarded as density reward. It will highly pump the acquisition value in the area that has fewer observed points in dataset $D_i$, and encourage the solver to search this sparse area. 

Equipped with the density-based adjustment strategy, observed sample points in $D_{i}$ will distribute more evenly to enhance the efficiency of the neighbor-based regularization.
Note that in practice, $\sigma_2$ also moves according to the searching progress and we will discuss it in section~\ref{subsec:3.2.3}.

%--------------------------------------------------------------------------------------------------------------------------------------------------
\subsubsection{\textbf{Dynamic regularization strength and density reward}}\label{subsec:3.2.3}
In Eq \ref{eq3}, we adopt a fixed $\sigma_1$ to smooth the observations and regularize the surrogate model. In practice, we use a dynamic $\sigma_1$ to weaken the strength of the regularization as the search progresses. 

At the beginning of the search process, we adopt a larger $\sigma_1$ to strengthen the smoothness. This helps the model to fit the observation dataset with lower resolution and concentrate more on long-range trends in the entire search space. As the search progresses, the $\sigma_1$ will gradually decrease to weaken the regularization, encouraging the model to fit more details when the solver is more close to the optimal solution. Specifically, we can rewrite the filter function in equation \ref{eq3} as follows:
\begin{equation}
    P_j(\boldsymbol{x}_k) = \left\{ 
    \begin{array}{ll}
    1 & \textrm{\qquad$\Vert \boldsymbol{x}_j - \boldsymbol{x}_k \Vert \leq \sigma_1(i) $}\\
    0 & \textrm{\qquad$\Vert \boldsymbol{x}_j - \boldsymbol{x}_k \Vert > \sigma_1(i) $}
    \end{array} 
    \right.
    \quad \sigma_1(i) = \sigma_{1}^{0} + (1-i/N)*\sigma_{1}^{1}
\end{equation}
where $\sigma_{1}^{0}$ is the base regularization strength, $i$ is the current optimization iteration and $N$ is the total number of the optimization iteration. The neighbor radius will starts at $\sigma_{1}^{0} + \sigma_{1}^{1}$ and end up in $\sigma_{1}^{0}$.

Similarly, we also adopt a dynamic $\sigma_2$ in Eq \ref{eq5}, the difference is that it moves in an opposite direction. At the beginning of the search process, we use a relatively smaller $\sigma_2$ to encourage the algorithm to search the entire search space especially the sparse area that contains fewer observed points. Then the $\sigma_2$ will gradually decrease the density reward and rebalance the exploration-exploitation trade-off. Formally, we re-write the equation \ref{eq5} as
\begin{equation}
    g_d(\boldsymbol{x}) = e^{-f_n(\boldsymbol{x},\sigma_2(i))}, 
    \quad \sigma_2(i) = \sigma_{2}^{0} + i/N*\sigma_{2}^{1}
\end{equation}

It helps the search process to get rid of the local optimal at very beginning and finally rebalance the exploration and exploitation to the normal state because the adjustment item value $g_d$ will finally tend to be zero throughout the entire search space, making the searching process easier to converge on the optimal point.

\section{Experiments}\label{sec:exp}

We conduct extensive experiments on both conventional BO benchmark bayesmark and some commonly used computer vision benchmarks such as ImageNet~\cite{krizhevsky2012imagenet}, COCO~\cite{lin2014microsoft}. 
We first compare NRBO against several state-of-the-arts BO methods on both bayesmark (section~\ref{subsec:main_bayes}) and computer vision benchmarks (section~\ref{subsec:main_cv}). 
Then we conduct a more detailed ablation study to demonstrate the impact of each component of NRBO in section~\ref{subsec:ablation}. 
In addition, we present some qualitative visualization of NRBO in section~\ref{subsec:visualization}.
%

% \begin{figure*}[!tbp]
% \centering
%     % \begin{center}
%        \includegraphics[width=1.0\textwidth]{images/convergence.png}
%     % \end{center}
%    \caption{The convergence progress of different methods. The results indicate that NRBO converges faster and reaches better score on both bayesmarks and CV tasks. For VOC task, score represents the AP50 metric.}
%    \label{fig:fastab}
% \end{figure*}

% \begin{figure}[h!]
% \centering
%     % \begin{center}
%        \includegraphics[width=0.5\textwidth]{images/all_9_methods.png}
%     % \end{center}
%    \caption{The average performance of all 108 tasks in bayesmark benchmark.}
%    \label{fig:all_9_methods}
% \end{figure}

%--------------------------------------------------------------------------------------------------------------------------------------------------
\begin{table*}[t]
\begin{center}
\footnotesize
\setlength\tabcolsep{1pt}
\begin{tabular}{lccccccc|c c}
\toprule
Method & MLP\_digits & RF\_breast & SVM\_digits & SVM\_wine & ada\_breast &ada\_digits & linear\_breast & $Avg.\_score$ & $Avg.\_norm$\\
\midrule
Random & 98.63 & 92.07 & 94.95 & 81.19  & 93.75 & 69.26 & 89.27 &  90.93 & 1.07 \\
Hyperopt & 100.29 & 96.50 & 96.87 & 88.72 & 98.96 & 73.87 & 92.85 & 96.53 & 0.411  \\
Opentuner & 100.76 & 92.07 & 98.43 & 92.61 & 98.96 & 100.21 & 94.39 & 93.16 & 0.81 \\
Nevergrad & 100.11  & 98.83 & 100.07 & 70.30 & 98.96 & 76.09 & 94.55 & 93.95 & 0.72 \\
Pysot & 101.13 & 95.34 & 97.56 & 99.87 & 95.83 & 96.55 & 97.80 & 98.34 & 0.20 \\
Skopt & 100.08 & 98.60 & 96.86 & 96.24 & 93.75 & 76.79 & 89.88 & 96.23 & 0.45 \\
Turbo & 101.10 & 96.74 & 98.53 & 100 & 95.83 & 91.45 & 97.56 & 98.28 & 0.20 \\
HEBO & 101.38 & 97.67 & 101.70 & 96.24 & 98.96 & 112.88 & 95.82 & 99.94 & 0.01 \\
NRBO & 101.49 & 100.00 & 103.53 & 103.63 & 100.00 & 113.01 & 101.01 & 100.34 & -0.04 \\
\bottomrule
\end{tabular}
\caption{\textbf{The performance of 9 optimizers on bayesmark benchmark.} We list the average loss score(denoted as Avg.score) and normalized mean score(denoted as Avg.norm) for all 108 tasks in bayesmark benchmark.}
\label{tab:main_bayes}
\end{center}
\end{table*}
%--------------------------------------------------------------------------------------------------------------------------------------------------

\subsection{Results on Bayesmark}\label{subsec:main_bayes}

Bayesmark\footnote{https://github.com/uber/bayesmark} benchmark contains 6 standard datasets (breast, digits, iris, wine, boston, diabetes) and 9 commonly used machine learning models (DT, MLP-ADAM, MLP-SGD, RF, SVM, ADA, KNN, Lasso, Linear). Each model has two variants for classification and regression. 
Combining these datasets and models, there are 108 tasks to evaluate the hyperparameter optimization algorithms. 
%
% On the Bayesmark, we set the number of iterations to 16 and the batch size to 8 for all tasks, and each optimization task repeats 3 times. We set $\sigma_{1}^{0}$, $\sigma_{2}^{0}$ to 0, $\sigma_{1}^{1}$ to 0.15 and  $\sigma_{2}^{1}$ to 1.0 for all experiments. Similar to HEBO, we adopt a sobel sampling initialization at the first four iterations and we follow the standard hyperparameter space setting for each task.

\begin{figure*}[t]
\centering
    \begin{center}
    \includegraphics[width=1.0\textwidth]{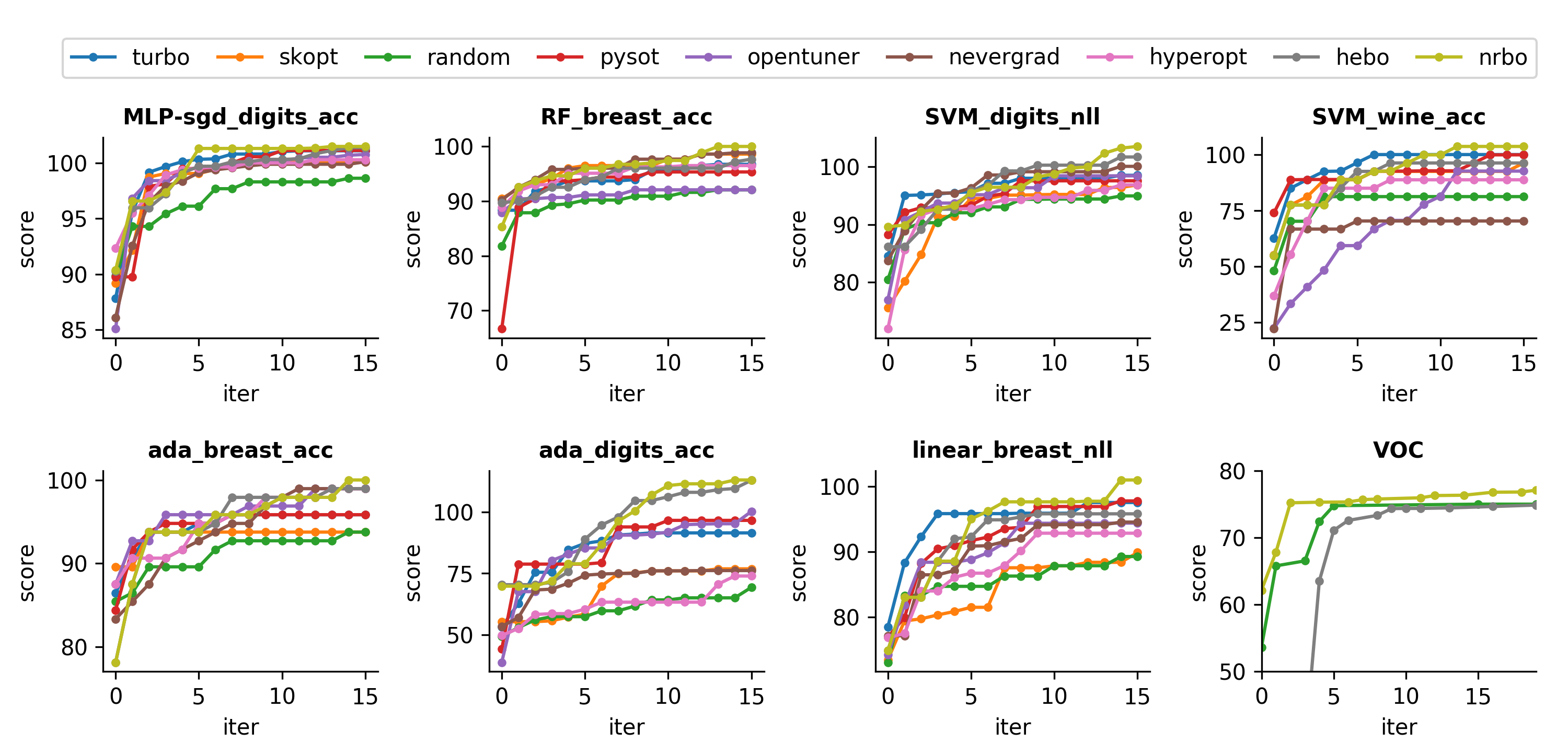}
    \end{center}
    \caption{The convergence progress of different methods. The results indicate that NRBO converges faster and reaches better score on both bayesmarks and CV tasks. For VOC task, score represents the AP50 metric.}
    \label{fig:fastab}
\end{figure*}

We adopt two official metrics to measure the performance on Bayesmark. The loss score metric for each task is calculated by $100 * (1-loss)$ . The normalized mean score first calculates the performance gap between observations and the global optimal point, then normalize it by the gap between random search results and the optimal.
Table \ref{tab:main_bayes} shows the performance of NRBO and other 8 commonly used hyperparameter optimizers, including random search, hyperopt~\cite{bergstra2013making}, opentuner~\cite{ansel2014opentuner}, pysot~\cite{eriksson2019pysot}, skopt~\cite{Timgates422020Scikit-optimize}, turbo~\cite{eriksson2019scalable}, nevergrad~\cite{nevergrad} and HEBO~\cite{cowen2020hebo}.  
We also select 7 tasks in bayesmark and plot the optimization process in detail at figure \ref{fig:fastab}. NRBO starts to surpass other optimizers at the 9-th iteration and stays ahead till the end of the optimization. The results indicate that NRBO converges faster and reaches better final score.
% We also select 7 tasks in bayesmark and plot the optimization process in detail at figure \ref{fig:fastab}. The results indicate that NRBO converges faster and reaches better final score.
% %
% Figure \ref{fig:all_9_methods} lists the normalized mean scores of each optimizer at different search iterations.  It demonstrates the performance gap with different computation resources. NRBO starts to surpass other optimizers at the 9-th iteration and stays ahead till the end of the optimization. We also select 7 tasks in bayesmark and plot the optimization process in detail at figure \ref{fig:fastab}. The results indicate that NRBO converges faster and reaches better final score.

%--------------------------------------------------------------------------------------------------------------------------------------------------
%--------------------------------------------------------------------------------------------------------------------------------------------------

\subsection{Results on Computer Vision Tasks}\label{subsec:main_cv}
In this section, we show the experiment results about different hyperparameter optimization algorithms used on common computer vision tasks. We only choose three optimization algorithms including random search, HEBO and our NRBO due to the high cost of computer vision tasks on both time and resource.

Our computer vision experiments include classification task and detection task.
% Table~\ref{tab:cv_space} presents the hyperparameter space and the corresponded bound for both classification and detection tasks.
%
For the classification task, we use ResNet-18~\cite{he2016deep} and train it on ImageNet~\cite{krizhevsky2012imagenet}, CIFAR10~\cite{krizhevsky2009learning}, CIFAR100~\cite{krizhevsky2009learning} and Stanford Cars~\cite{KrauseStarkDengFei-Fei_3DRR2013} datasets. Four hyperparameters including momentum, weight decay, label smooth and learning rate compose a 4-dimension searching space. 
% 
% To reduce the computation cost, we set a proxy task when training on ImageNet, the learning rate drops at the 20,000th, 25,000th, 27,500th iteration respectively with the factor of 10x, and the total iteration number is set to 30,000. On other vision classification tasks, the total training epoch is set to 200 and no proxy task is used.
%
For the detection task, we use RetinaNet~\cite{lin2017focal} and train it on Pascal VOC~\cite{Everingham10} and MS-COCO2017~\cite{lin2014microsoft} datasets. The searching space is also composed of 4 hyperparameters including momentum, weight decay, positive IoU thresh and negative IoU thresh. 
%
% Weight decay and learning rate are mapped into a log space during searching procedure in both tasks. The hyperparameter optimization experiment of each task repeats 4 times in order to make results stable and convincing. 
%
% Please note that, like ImageNet task, we shrink the training scheduler as a proxy to reduce the search cost when training on COCO. The learning rate drops at the 5th and 7th epoch respectively and the total epoch number is set to 8. 
% The full-training results on COCO and ImageNet will be shown in section~\ref{subsec:fulltrain}.

The best detection results with different hyperparameter optimization algorithms are shown in table~\ref{tab:main_cv}, where HEBO performs better than random search on COCO, and NRBO performs best on both VOC and COCO datasets. These results are consistent with those in bayesmark benchmark, demonstrating the effectiveness of our proposed method. 
% We also present a more detailed optimization process on VOC dataset in figure \ref{fig:fastab}.

%--------------------------------------------------------------------------------------------------------------------------------------------------
\begin{table*}[t]
\begin{center}
\small
\footnotesize
\setlength\tabcolsep{3pt}
\begin{tabular}{lcccccc}
\toprule
Method & ImageNet & VOC &CIFAR10 &CIFAR100 & Stanford Car  & COCO   \\
\midrule
Random & 62.00    &  75.02 & 95.28 & 81.96 & 87.09 & 31.56 \\
HEBO &  62.07(+0.07)  & 74.88(-0.14) & 95.35(+0.07)  & 81.91(-0.05) & 87.56(+0.47) &  32.34(+0.78)   \\
NRBO &  62.22(+0.22) & 77.12(+2.10) &  95.43(+0.15) & 82.16(+0.20) & 87.94(+0.85) &  33.92(+1.36) \\
\bottomrule
\end{tabular}
\caption{Experiment results on computer vision tasks. The positive numbers in parentheses represent absolute improvements relative to Random Search.}
\label{tab:main_cv}
\end{center}
\end{table*}
%--------------------------------------------------------------------------------------------------------------------------------------------------

The classification results are shown in table~\ref{tab:main_cv}. It is seen that the performance of HEBO is close to the naive random search. This may be caused by a relatively flat performance landscape near the optimal solution in search space, and even a random search could have a high probability of getting a good result.
%This may be because that the optimal solution area in this task are so large that the random search has a strong possibility to find out the relatively optimal solution area, and 
Compared to bayesmark and COCO, the output of the ImageNet experiments with a specific hyperparameters is more stable. Therefore, the NRBO could hardly benefit from the regularization. Even in this case, NRBO still surpasses the HEBO on all four classification tasks.
%Even in the case that the average results of HEBO is only 0.064 higher than Random, our NRBO is 0.161 higher than HEBO, maintaining a robust performance.

\subsection{Full-training Results of Computer Vision Tasks} \label{subsec:ablation}

Since the hyperparameter optimization needs repetitive trials, the computation cost is unaffordable on large-scale datasets. To overcome the computation resource limitation, in large computer vision datasets ImageNet and COCO, we optimize the hyperparameter on proxy tasks with \textit{60\%} of total iterations for only 20 trials. Then we use the hyperparameters optimized on the \textit{\textbf{proxy}} task for a \textit{\textbf{full}} training to verify whether the NRBO is still in the lead. We repeat each optimizer 4 times for stable and convincing results.

\textbf{Classification} The configuration is same as ImageNet experiments in section~\ref{subsec:main_cv}, except that the learning rate of proxy task decreases at 25000th, 50000th, and 75000th iteration and ends at 80000th iteration. In full-training task, the learning rate schedule time is set as 37500, 75000, and 112500. Training ends at the 125000th iteration. As shown in table~\ref{tab:full_exp}, the average accuracy of NRBO surpasses HEBO and random search by 0.37 and 0.33 respectively in the full-training setting.

\textbf{Detection} The configuration is same as COCO experiments in section~\ref{subsec:main_cv}. In full-training task, the learning rate drops at the 9th and 12th epoch. Training ends at the 14th epoch. As shown in table~\ref{tab:full_exp}, the mAP of NRBO surpasses HEBO and random search by 2.29 and 1.37 respectively in the full-training setting.

Experiments above demonstrate that the hyperparameters searched by the small proxy task can be transfered to the full-training task. 

\begin{table*}[h!]
% ---------------------------------------------------------------
\begin{minipage}[t]{0.5\textwidth}
\centering
\setlength{\tabcolsep}{4.0pt}
\footnotesize
\begin{tabular}{ccccc}
\toprule
Exp. Index            & Task  & Random & HEBO   & NRBO \\
\midrule
\multirow{2}{*}{1}    & proxy & 67.268 & 65.808 & 66.490 \\
                      & full  & 68.324 & 67.660 & 68.362 \\
\midrule
\multirow{2}{*}{2}    & proxy & 65.644 & 67.258 & 66.972 \\
                      & full  & 67.858 & 68.642 & 68.562 \\
\midrule
\multirow{2}{*}{3}    & proxy & 66.290 & 66.908 & 67.058 \\
                      & full  & 67.668 & 68.494 & 68.446 \\
\midrule
\multirow{2}{*}{4}    & proxy & 66.916 & 66.116 & 67.020 \\
                      & full  & 68.638 & 67.848 & 68.606 \\
\midrule
\multirow{2}{*}{Avg.} & proxy & 66.530 & 66.523 & \textbf{66.885} \\
                      & full  & 68.122 & 68.161 & \textbf{68.494} \\
\bottomrule
\end{tabular}
\end{minipage}
% ---------------------------------------------------------------
\begin{minipage}[t]{0.5\textwidth}
\centering
\setlength{\tabcolsep}{4.0pt}
\footnotesize
\begin{tabular}{ccccc}
\toprule
Exp. Index            & Task  & Random & HEBO   & NRBO \\
\midrule
\multirow{2}{*}{1}    & proxy & 32.283 & 32.070 & 34.853 \\
                      & full  & 34.549 & 34.438 & 36.952 \\
\midrule
\multirow{2}{*}{2}    & proxy & 32.085 & 32.334 & 33.231 \\
                      & full  & 33.773 & 34.609 & 35.634 \\
\midrule
\multirow{2}{*}{3}    & proxy & 31.705 & 33.571 & 34.410 \\
                      & full  & 34.123 & 35.997 & 36.199 \\
\midrule
\multirow{2}{*}{4}    & proxy & 30.147 & 31.395 & 33.222 \\
                      & full  & 32.487 & 33.585 & 35.307 \\
\midrule
\multirow{2}{*}{Avg.} & proxy & 31.555 & 32.343 & \textbf{33.929} \\
                      & full  & 33.733 & 34.657 & \textbf{36.023} \\
\bottomrule
\end{tabular}
\end{minipage}
% ---------------------------------------------------------------
\caption{(Left) Proxy task and full-training task results on ImageNet. (Right) Proxy task and full-training task results on COCO.}
\label{tab:full_exp}
\end{table*}

%--------------------------------------------------------------------------------------------------------------------------------------------------
\subsection{Ablation Study} \label{subsec:ablation}
% As shown in Fig~\ref{fig:fastab}, 

In this section, we analyze the efficiency of each component proposed in NRBO. We first conduct experiments with a variant NRBO that cancels the density-based acquisition function. If we further cancel the regularization mechanism proposed in section~\ref{subsec:3.2.1}, the NRBO will degenerate to a normal HEBO optimizer.

\begin{figure}[htbp]
\centering
% ---------------------------------------------------------------
\begin{minipage}[t]{0.45\textwidth}
\centering
\includegraphics[width=0.9\textwidth]{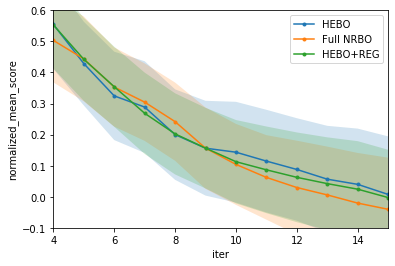}
\end{minipage}
% ---------------------------------------------------------------
\begin{minipage}[t]{0.45\textwidth}
\centering
\includegraphics[width=0.9\textwidth]{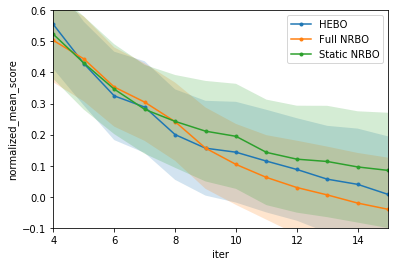}
\end{minipage}
% ---------------------------------------------------------------
\caption{\textbf{(Left) Ablation study for regularization and acquisition function}. HEBO denotes the NRBO w/o regularization and density-based acquisition reward. HEBO+REG denotes the NRBO w/o density-based acquisition reward. \textbf{(Right) Ablation study for dynamic regularization and acquisition function}. Static NRBO denotes the NRBO w/o dynamic regularization and acquisition function.}
\label{fig:abla}
% ---------------------------------------------------------------
\end{figure}

Figure~\ref{fig:abla} left shows the performance of HEBO, variant NRBO that without density-based acquisition function and a full NRBO. We can find that the variant NRBO still outperforms the basic HEBO within a wide range of computation resource settings. The variant NRBO significantly surpasses the HEBO from the tenth to the twelfth iteration. As the optimization progresses, the variant NRBO and HEBO finally reach the tie in the end. Note that the full NRBO still outperforms the HEBO at the end of the optimization. It suggests that the density-based acquisition function helps the optimizer jump out of the local optimal and leads to a better solution.

We also compare the NRBO with its variant that cancels the dynamic regularization strength and density reward described in section~\ref{subsec:3.2.3}. Figure~\ref{fig:abla} right shows that, without the dynamic regularization strength and density reward, the performance of NRBO deteriorates severely as the optimization progresses. This is reasonable because always keeping a strong regularization will prevent the surrogate model from fitting the subtle patterns at the high-performance area where the search process is close to optimal. In addition, a fixed density-based acquisition reward encourages the optimizer to explore rather than exploit. It will also prevent the optimizer from a detailed search in the high-performance area.

\subsection{Qualitative Visualization}\label{subsec:visualization}

In this section we visualize the effectiveness of regularization on an example dataset.  We compare the standard Gaussian process Bayesian optimizer with a neighbor-based regularized optimizer on different noise levels. The search space is normalized to $[0,1]$ and the neighbor radius for regularization is set to 0.1. The example dataset is generated with $sin(2\pi x) + cos(2\pi y)+\epsilon\sigma$, where $\epsilon$ is the noise level and $\sigma \sim N(0,1)$. 

\begin{figure}
\centering
\includegraphics[width=0.9\textwidth]{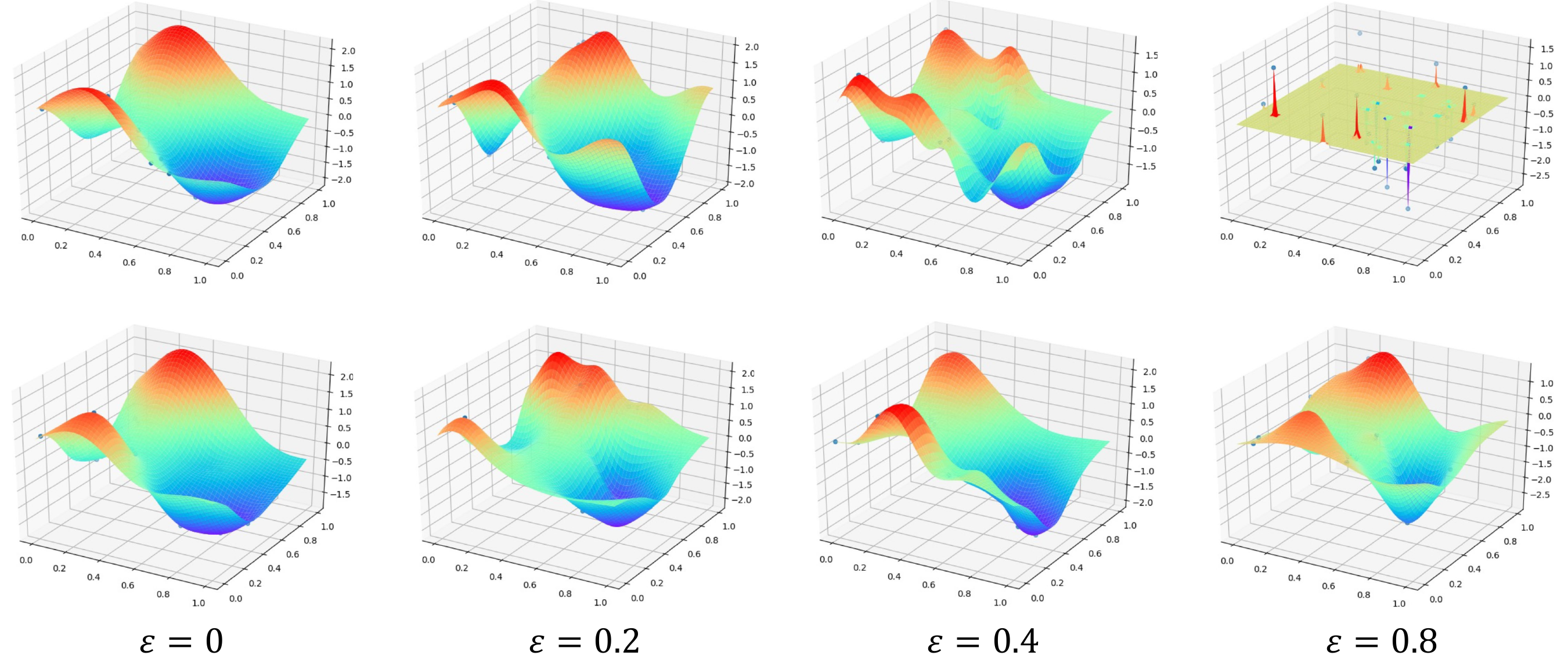}
\caption{\textbf{Fitting surrogate model on the same dataset with different nosise level.} From left to right: Gradually increased noise level $\epsilon$, from $0$ to $0.8$. Top row: A typical Bayesian optimization with Gaussian process and Matern Kernel. Bottom row: Neighbor regularized Bayesian optimization.}
\label{fig:noise}
\end{figure}

As shown in Figure~\ref{fig:noise}, in the first row, both of them fit the dataset well when the noise level is 0. When the noise level is set to 0.4, standard Gaussian process model starts to overfit the noise data and its output tends to be sharp and unstable. If we further increase the noise level to 0.8, the standard Gaussian process model completely collapsed. Meanwhile, the regularized model still outputs a smooth prediction that is very close to the ground truth.

\section{Conclusion}
We propose a novel hyperparameter optimization algorithm NRBO in this work. 
We first propose a neighbor-based regularization to smooth sample observations by the neighbor statistics. 
To further improve the stability of the neighbor statistics, we propose a density-based acquisition function.
In addition, an adjustment mechanism is adopted to adjust the regularization strength and acquisition reward based on the remaining computation resources.
Extensive experiments demonstrate NRBO could accelerate the convergence of hyperparameter optimization and reduce the risk of collapse. 
%
% NRBO achieves the state-of-the-art performance on both the bayesmark and important computer vision benchmarks such as CIFAR10/100, StanfordCars, ImageNet, Pascal VOC and COCO.
%

%\noindent\textbf{Limitations}. We propose a neighbor-based regularization paradigm to accelerate and regularize the hyperparameter optimization under the Bayesian optimization framework. 
%
%However, the computation cost is still high in heavy tasks such as ImageNet and COCO. 
%
%Determine how to design a more lightweight and efficient proxy task or model is crucial for these heavy tasks.
%
%We leave them as future works to explore.

\bibliography{egbib}

\begin{thebibliography}{35}
\providecommand{\natexlab}[1]{#1}
\providecommand{\url}[1]{\texttt{#1}}
\expandafter\ifx\csname urlstyle\endcsname\relax
  \providecommand{\doi}[1]{doi: #1}\else
  \providecommand{\doi}{doi: \begingroup \urlstyle{rm}\Url}\fi

\bibitem[Ansel et~al.(2014)Ansel, Kamil, Veeramachaneni, Ragan-Kelley, Bosboom,
  O'Reilly, and Amarasinghe]{ansel2014opentuner}
Jason Ansel, Shoaib Kamil, Kalyan Veeramachaneni, Jonathan Ragan-Kelley,
  Jeffrey Bosboom, Una-May O'Reilly, and Saman Amarasinghe.
\newblock Opentuner: An extensible framework for program autotuning.
\newblock In \emph{Proceedings of the 23rd international conference on Parallel
  architectures and compilation}, pages 303--316, 2014.

\bibitem[Bergstra and Bengio(2012)]{bergstra2012random}
James Bergstra and Yoshua Bengio.
\newblock Random search for hyper-parameter optimization.
\newblock \emph{Journal of machine learning research}, 13\penalty0 (2), 2012.

\bibitem[Bergstra et~al.(2013)Bergstra, Yamins, and Cox]{bergstra2013making}
James Bergstra, Daniel Yamins, and David Cox.
\newblock Making a science of model search: Hyperparameter optimization in
  hundreds of dimensions for vision architectures.
\newblock In \emph{International conference on machine learning}, pages
  115--123. PMLR, 2013.

\bibitem[Cowen-Rivers et~al.(2020)Cowen-Rivers, Lyu, Wang, Tutunov, Jianye,
  Wang, and Ammar]{cowen2020hebo}
Alexander~I Cowen-Rivers, Wenlong Lyu, Zhi Wang, Rasul Tutunov, Hao Jianye, Jun
  Wang, and Haitham~Bou Ammar.
\newblock Hebo: Heteroscedastic evolutionary bayesian optimisation.
\newblock \emph{arXiv e-prints}, pages arXiv--2012, 2020.

\bibitem[Daulton et~al.(2021)Daulton, Eriksson, Balandat, and Bakshy]{2021MOBO}
Samuel Daulton, David Eriksson, Maximilian Balandat, and Eytan Bakshy.
\newblock Multi-objective bayesian optimization over high-dimensional search
  spaces.
\newblock \emph{CoRR}, abs/2109.10964, 2021.
\newblock URL \url{https://arxiv.org/abs/2109.10964}.

\bibitem[Eggensperger et~al.(2021)Eggensperger, M{\"{u}}ller, Mallik, Feurer,
  Sass, Klein, Awad, Lindauer, and Hutter]{2021hpobench}
Katharina Eggensperger, Philipp M{\"{u}}ller, Neeratyoy Mallik, Matthias
  Feurer, Ren{\'{e}} Sass, Aaron Klein, Noor~H. Awad, Marius Lindauer, and
  Frank Hutter.
\newblock Hpobench: {A} collection of reproducible multi-fidelity benchmark
  problems for {HPO}.
\newblock In Joaquin Vanschoren and Sai{-}Kit Yeung, editors, \emph{Proceedings
  of the Neural Information Processing Systems Track on Datasets and Benchmarks
  1, NeurIPS Datasets and Benchmarks 2021, December 2021, virtual}, 2021.

\bibitem[Eriksson et~al.(2019{\natexlab{a}})Eriksson, Bindel, and
  Shoemaker]{eriksson2019pysot}
David Eriksson, David Bindel, and Christine~A Shoemaker.
\newblock pysot and poap: An event-driven asynchronous framework for surrogate
  optimization.
\newblock \emph{arXiv preprint arXiv:1908.00420}, 2019{\natexlab{a}}.

\bibitem[Eriksson et~al.(2019{\natexlab{b}})Eriksson, Pearce, Gardner, Turner,
  and Poloczek]{eriksson2019scalable}
David Eriksson, Michael Pearce, Jacob Gardner, Ryan~D Turner, and Matthias
  Poloczek.
\newblock Scalable global optimization via local bayesian optimization.
\newblock \emph{Advances in Neural Information Processing Systems},
  32:\penalty0 5496--5507, 2019{\natexlab{b}}.

\bibitem[Everingham et~al.(2010)Everingham, Van~Gool, Williams, Winn, and
  Zisserman]{Everingham10}
M.~Everingham, L.~Van~Gool, C.~K.~I. Williams, J.~Winn, and A.~Zisserman.
\newblock The pascal visual object classes (voc) challenge.
\newblock \emph{International Journal of Computer Vision}, 88\penalty0
  (2):\penalty0 303--338, June 2010.

\bibitem[Falkner et~al.(2018)Falkner, Klein, and Hutter]{falkner2018bohb}
Stefan Falkner, Aaron Klein, and Frank Hutter.
\newblock Bohb: Robust and efficient hyperparameter optimization at scale.
\newblock In \emph{International Conference on Machine Learning}, pages
  1437--1446. PMLR, 2018.

\bibitem[Fix and Hodges(1989)]{fix1989knn}
Evelyn Fix and Joseph~Lawson Hodges.
\newblock Discriminatory analysis. nonparametric discrimination: Consistency
  properties.
\newblock \emph{International Statistical Review/Revue Internationale de
  Statistique}, 57\penalty0 (3):\penalty0 238--247, 1989.

\bibitem[Golovin et~al.(2017)Golovin, Solnik, Moitra, Kochanski, Karro, and
  Sculley]{golovin2017google}
Daniel Golovin, Benjamin Solnik, Subhodeep Moitra, Greg Kochanski, John Karro,
  and David Sculley.
\newblock Google vizier: A service for black-box optimization.
\newblock In \emph{Proceedings of the 23rd ACM SIGKDD international conference
  on knowledge discovery and data mining}, pages 1487--1495, 2017.

\bibitem[Hazan et~al.(2017)Hazan, Klivans, and Yuan]{hazan2017hyperparameter}
Elad Hazan, Adam Klivans, and Yang Yuan.
\newblock Hyperparameter optimization: A spectral approach.
\newblock \emph{arXiv preprint arXiv:1706.00764}, 2017.

\bibitem[He et~al.(2016)He, Zhang, Ren, and Sun]{he2016deep}
Kaiming He, Xiangyu Zhang, Shaoqing Ren, and Jian Sun.
\newblock Deep residual learning for image recognition.
\newblock In \emph{Proceedings of the IEEE conference on computer vision and
  pattern recognition}, pages 770--778, 2016.

\bibitem[Hutter et~al.(2011)Hutter, Hoos, and
  Leyton-Brown]{hutter2011sequential}
Frank Hutter, Holger~H Hoos, and Kevin Leyton-Brown.
\newblock Sequential model-based optimization for general algorithm
  configuration.
\newblock In \emph{International conference on learning and intelligent
  optimization}, pages 507--523. Springer, 2011.

\bibitem[Krause et~al.(2013)Krause, Stark, Deng, and
  Fei-Fei]{KrauseStarkDengFei-Fei_3DRR2013}
Jonathan Krause, Michael Stark, Jia Deng, and Li~Fei-Fei.
\newblock 3d object representations for fine-grained categorization.
\newblock In \emph{4th International IEEE Workshop on 3D Representation and
  Recognition (3dRR-13)}, Sydney, Australia, 2013.

\bibitem[Krizhevsky et~al.(2009)Krizhevsky, Hinton,
  et~al.]{krizhevsky2009learning}
Alex Krizhevsky, Geoffrey Hinton, et~al.
\newblock Learning multiple layers of features from tiny images.
\newblock 2009.

\bibitem[Krizhevsky et~al.(2012)Krizhevsky, Sutskever, and
  Hinton]{krizhevsky2012imagenet}
Alex Krizhevsky, Ilya Sutskever, and Geoffrey~E Hinton.
\newblock Imagenet classification with deep convolutional neural networks.
\newblock \emph{Advances in neural information processing systems},
  25:\penalty0 1097--1105, 2012.

\bibitem[Lerman(1980)]{lerman1980fitting}
PM~Lerman.
\newblock Fitting segmented regression models by grid search.
\newblock \emph{Journal of the Royal Statistical Society: Series C (Applied
  Statistics)}, 29\penalty0 (1):\penalty0 77--84, 1980.

\bibitem[Li et~al.(2017)Li, Jamieson, DeSalvo, Rostamizadeh, and
  Talwalkar]{li2017hyperband}
Lisha Li, Kevin Jamieson, Giulia DeSalvo, Afshin Rostamizadeh, and Ameet
  Talwalkar.
\newblock Hyperband: A novel bandit-based approach to hyperparameter
  optimization.
\newblock \emph{The Journal of Machine Learning Research}, 18\penalty0
  (1):\penalty0 6765--6816, 2017.

\bibitem[Liashchynskyi and Liashchynskyi(2019)]{liashchynskyi2019grid}
Petro Liashchynskyi and Pavlo Liashchynskyi.
\newblock Grid search, random search, genetic algorithm: A big comparison for
  nas.
\newblock \emph{arXiv preprint arXiv:1912.06059}, 2019.

\bibitem[Lin et~al.(2014)Lin, Maire, Belongie, Hays, Perona, Ramanan,
  Doll{\'a}r, and Zitnick]{lin2014microsoft}
Tsung-Yi Lin, Michael Maire, Serge Belongie, James Hays, Pietro Perona, Deva
  Ramanan, Piotr Doll{\'a}r, and C~Lawrence Zitnick.
\newblock Microsoft coco: Common objects in context.
\newblock In \emph{European conference on computer vision}, pages 740--755.
  Springer, 2014.

\bibitem[Lin et~al.(2017)Lin, Goyal, Girshick, He, and
  Doll{\'a}r]{lin2017focal}
Tsung-Yi Lin, Priya Goyal, Ross Girshick, Kaiming He, and Piotr Doll{\'a}r.
\newblock Focal loss for dense object detection.
\newblock In \emph{Proceedings of the IEEE international conference on computer
  vision}, pages 2980--2988, 2017.

\bibitem[Lindauer et~al.(2022)Lindauer, Eggensperger, Feurer, Biedenkapp, Deng,
  Benjamins, Ruhkopf, Sass, and Hutter]{2022smac3}
Marius Lindauer, Katharina Eggensperger, Matthias Feurer, Andr{\'{e}}
  Biedenkapp, Difan Deng, Carolin Benjamins, Tim Ruhkopf, Ren{\'{e}} Sass, and
  Frank Hutter.
\newblock {SMAC3:} {A} versatile bayesian optimization package for
  hyperparameter optimization.
\newblock \emph{J. Mach. Learn. Res.}, 23:\penalty0 54:1--54:9, 2022.

\bibitem[Rapin and Teytaud(2018)]{nevergrad}
J.~Rapin and O.~Teytaud.
\newblock {Nevergrad - A gradient-free optimization platform}.
\newblock \url{https://GitHub.com/FacebookResearch/Nevergrad}, 2018.

\bibitem[Rasmussen(2003)]{rasmussen2003gaussian}
Carl~Edward Rasmussen.
\newblock Gaussian processes in machine learning.
\newblock In \emph{Summer school on machine learning}, pages 63--71. Springer,
  2003.

\bibitem[Shahriari et~al.(2015)Shahriari, Swersky, Wang, Adams, and
  De~Freitas]{shahriari2015taking}
Bobak Shahriari, Kevin Swersky, Ziyu Wang, Ryan~P Adams, and Nando De~Freitas.
\newblock Taking the human out of the loop: A review of bayesian optimization.
\newblock \emph{Proceedings of the IEEE}, 104\penalty0 (1):\penalty0 148--175,
  2015.

\bibitem[Snoek et~al.(2012)Snoek, Larochelle, and Adams]{snoek2012practical}
Jasper Snoek, Hugo Larochelle, and Ryan~P Adams.
\newblock Practical bayesian optimization of machine learning algorithms.
\newblock \emph{Advances in neural information processing systems}, 25, 2012.

\bibitem[Snoek et~al.(2015)Snoek, Rippel, Swersky, Kiros, Satish, Sundaram,
  Patwary, Prabhat, and Adams]{snoek2015scalable}
Jasper Snoek, Oren Rippel, Kevin Swersky, Ryan Kiros, Nadathur Satish,
  Narayanan Sundaram, Mostofa Patwary, Mr~Prabhat, and Ryan Adams.
\newblock Scalable bayesian optimization using deep neural networks.
\newblock In \emph{International conference on machine learning}, pages
  2171--2180. PMLR, 2015.

\bibitem[Solis and Wets(1981)]{solis1981minimization}
Francisco~J Solis and Roger J-B Wets.
\newblock Minimization by random search techniques.
\newblock \emph{Mathematics of operations research}, 6\penalty0 (1):\penalty0
  19--30, 1981.

\bibitem[Swersky et~al.(2013)Swersky, Snoek, and Adams]{swersky2013multi}
Kevin Swersky, Jasper Snoek, and Ryan~P Adams.
\newblock Multi-task bayesian optimization.
\newblock \emph{Advances in neural information processing systems}, 26, 2013.

\bibitem[{Timgates42}(2020)]{Timgates422020Scikit-optimize}
{Timgates42}.
\newblock {scikit-optimize}, 2020.

\bibitem[Victoria and Maragatham(2021)]{victoria2021automatic}
A~Helen Victoria and G~Maragatham.
\newblock Automatic tuning of hyperparameters using bayesian optimization.
\newblock \emph{Evolving Systems}, 12\penalty0 (1):\penalty0 217--223, 2021.

\bibitem[Wu et~al.(2019)Wu, Chen, Zhang, Xiong, Lei, and
  Deng]{wu2019hyperparameter}
Jia Wu, Xiu-Yun Chen, Hao Zhang, Li-Dong Xiong, Hang Lei, and Si-Hao Deng.
\newblock Hyperparameter optimization for machine learning models based on
  bayesian optimization.
\newblock \emph{Journal of Electronic Science and Technology}, 17\penalty0
  (1):\penalty0 26--40, 2019.

\bibitem[Yang and Shami(2020)]{yang2020hyperparameter}
Li~Yang and Abdallah Shami.
\newblock On hyperparameter optimization of machine learning algorithms: Theory
  and practice.
\newblock \emph{Neurocomputing}, 415:\penalty0 295--316, 2020.

\end{thebibliography}
\end{document}